\pdfoutput=1

\documentclass[11pt]{article}

\usepackage{EMNLP2022}

\usepackage{times}
\usepackage{latexsym}

\usepackage[T1]{fontenc}

\usepackage[utf8]{inputenc}

\usepackage{microtype}

\usepackage{inconsolata}
\usepackage{inconsolata}
\usepackage{subfigure}
\usepackage{microtype}
\usepackage{caption}
\usepackage{bm}
\usepackage{url}
\usepackage{graphicx}
\usepackage{multirow, booktabs}
\usepackage{amsmath}
\usepackage{enumitem}
\usepackage{amsfonts}
\usepackage{makecell}
\usepackage{verbatim}
\usepackage{array}
\usepackage{xcolor}
\usepackage{soul}

\title{Disentangling Task Relations for Few-shot Text Classification via Self-Supervised Hierarchical Task Clustering}
\author{Juan Zha*$^1$\thanks{~Most of the work was done when the first author was a research assistant at Southern University of Science and Technology; * Equal contribution; $^{\dagger}$Corresponding author.},
Zheng Li*$^2$, Ying Wei$^3$, Yu Zhang$^{\dagger}$$^4$\\
  $^1$University of Southern California, CA, USA \\
  $^2$Amazon.com Inc, CA, USA \\
  $^3$City University of Hong Kong, Hong Kong, China \\
  $^4$Southern University of Science and Technology, China\\
  $^1${juanzha@usc.com}, $^2${amzzhe@amazon.com} \\
  $^3${yingwei@cityu.edu.hk}, $^4${yu.zhang.ust@gmail.com}
}

\begin{document}
\maketitle
\begin{abstract}
Few-Shot Text Classification (FSTC) imitates humans to learn a new text classifier efficiently with only few examples, by leveraging prior knowledge from historical tasks. However, most prior works assume that all the tasks are sampled from a single data source, which cannot adapt to real-world scenarios where tasks are heterogeneous and lie in different distributions. As such, existing methods may suffer from their globally knowledge-shared mechanisms to handle the task heterogeneity. On the other hand, inherent task relation are not explicitly captured, making task knowledge unorganized and hard to transfer to new tasks. Thus, we explore a new FSTC setting where tasks can come from a diverse range of data sources. To address the task heterogeneity, we propose a self-supervised hierarchical task clustering (SS-HTC) method. SS-HTC not only customizes cluster-specific knowledge by dynamically organizing heterogeneous tasks into different clusters in hierarchical levels but also disentangles underlying relations between tasks to improve the interpretability. Extensive experiments on five public FSTC benchmark datasets demonstrate the effectiveness of SS-HTC.
\end{abstract}

\section{Introduction}
Recent advances in deep learning highly rely on massive human annotations. This reliance increases the burden of data collection and meanwhile hinders its potentials to the low-data regime, where the labeled data is scarce and difficult to obtain. Inspired by human beings' capabilities that can quickly learn with a few examples, Few-Shot Learning (FSL)~\cite{vinyals2016matching,finn2017model}, which aims to learn a classifier that generalizes well even with a few training instances per class, has recently attracted much attention.

In the NLP domain, Few-Shot Text Classification (FSTC)~\cite{han2018fewrel} has been
actively investigated in data-sparsity scenarios, e.g., relation classification~\cite{han2018fewrel}, event classification~\cite{deng2020meta}, and intent classification~\cite{zhang2020deep}, where new categories such as relations, events, or intent types tend to emerge frequently and lack sufficient annotations. Meta-learning (a.k.a. learning to learn) approaches~\cite{finn2017model}, 
which transfer prior knowledge from previous tasks to improve the effectiveness in learning new tasks, 
have achieved superior performance for FSTC~\cite{gao2019hybrid,sun2019hierarchical,bao2020fewshot}. The prior knowledge can be instantiated as a transferable metric space for retrieving nearest prototypes in~\cite{gao2019hybrid,sun2019hierarchical}, dynamic capsules in~\cite{geng2019induction}, and distributional signatures in~\cite{bao2020fewshot}, etc.

\begin{figure}[t]
\centering
\includegraphics[width=1.0\linewidth]{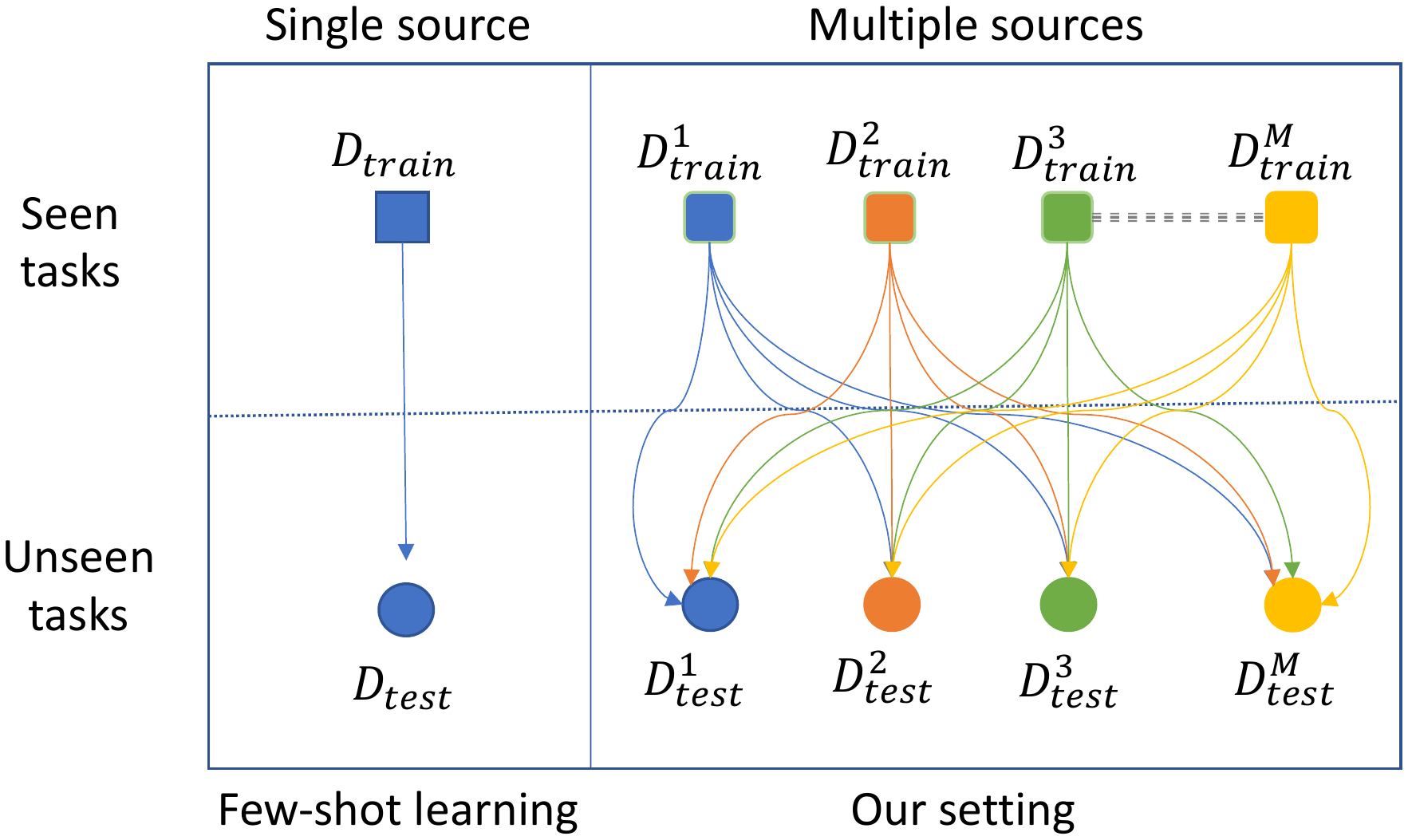}
\caption{Comparison of existing FSTC formulation and our proposed practical problem setting.}
\vspace{-5mm}
\label{fig:example}
\end{figure}

Despite their early success, such approaches have two main drawbacks: (1) they assume that all previous tasks are sampled from a single data source or domain, leading to tasks with low inter-task variance. 
As a consequence, these methods globally share the prior knowledge across all tasks but fail to handle real-world applications where the historical tasks that potentially contribute may come from diverse data sources in different distributions (a.k.a. task heterogeneity~\cite{vuorio2018toward}). For example, the knowledge learned from categorizing different types of products or services may be hardly transferred to classify public comments or opinions among different topics, or to determine users' intent in dialogues with chatbot services, with only limited labeled data; (2) handling heterogeneous tasks for the better generalization ability exactly requires reliable knowledge organization, which highly relies on disentangling underlying task relations that are ignored by prior works. Motivated by those, we study a new FSTC setting where tasks come from a diverse range of data sources with possibly different data distributions as shown in Figure~\ref{fig:example}. To embrace the skills learned from multiple task sources to improve the generalization ability, we propose a novel meta-learning framework named Self-Supervised Hierarchical Task Clustering (SS-HTC), which groups tasks into different clusters based on inherent task relations in multiple levels. When a new task arrives, it can quickly take advantage of the historical knowledge learned within the cluster it belongs to. 

Specifically, learning a superior task embedding is the cornerstone to disentangle underlying relations among tasks and group them into different clusters. However, a FSL task is hard to represent, as labeled training data in each FSL task are insufficient. To tackle this issue, we propose a label-oriented masked language modeling to recover the corresponding label texts using each training sample itself from the task. Such a self-supervised manner, considering informative label text semantics, encourages the model to generate more discriminative task embedding to discover reasonable task relationships even with limited label information.

After that, each task embedding is passed to a hierarchical task tree to dynamically perform soft task clustering in multiple levels, so that the knowledge is shared among highly related tasks in the same cluster but differentiated between different clusters of tasks. Then, the updated task embedding outputted by the task tree encodes the representation of the cluster it belongs to. This cluster representation is finally passed to modulate the prior knowledge, a metric space for finding nearest prototypes following~\cite{snell2017prototypical}, to be cluster-specific. In a nutshell, SS-HTC not only quickly accesses the most relevant cluster and tailors the prior knowledge to address the challenge of task heterogeneity, but also increases the model interpretability by disentangling task correlations. Empirically, extensive experiments on five public FSTC benchmark datasets demonstrate that SS-HTC significantly and consistently outperforms state-of-the-art FSTC methods by a large margin.

Our contributions can be summarized as follows.
(1) A more realistic FSTC setting that allows diverse tasks with different distributions is investigated; (2) A novel SS-HTC framework is proposed to both tackle task heterogeneity and improve the interpretability by hierarchical task clustering; (3) Extensive experiments verify the effectiveness of the proposed SS-HTC method.

\section{Preliminaries}

\noindent \textbf{Few-shot learning (FSL)}
\label{sec:fsl_definition}
Considering a task $\mathcal{T}=\{\mathcal{S}, \mathcal{Q}\}$ that contains the training set $\mathcal{S}$ and the testing set $\mathcal{Q}$, the objective of FSL is to learn a model $G$ for this task given only a few labeled samples in $\mathcal{S}$. Typically, FSL is characterized as a $N$-way $K$-shot problem with $\mathcal{S}$ containing $K$ labeled examples per class for $N$ classes, i.e., $\small{{ \mathcal{ S }\!=\!\{ ({ \mathbf{x} }^{j }_{ i },y_{ i })\} _{ i, j=1 }^{ N, K} }}$, where ${\mathbf{x} }^{j }_{ i }$ is the $j$-th sample for the $i$-th class $y_{ i }$. Then $\mathbf{x}^{\mathcal{Q}}$ denotes an unlabeled sample of $\mathcal{Q}$ belonging to one of the $N$ classes and $\hat{y}^{\mathcal{Q}}$ denotes the estimated label of a model, i.e., $G(\mathcal{S}, \mathbf{x}^{\mathcal{Q}})\!\rightarrow\!\hat{y}^{\mathcal{Q}}$. In many existing works on FSL, $\mathcal{S}$ and $\mathcal{Q}$ are also known as the \textit{support set} and \textit{query set}, respectively.

Traditional deep learning models would severely overfit on FSL tasks since only a few labeled samples cannot accurately represent the true data distribution, which will result in learning classifiers with high variance and generalizing poorly to new data. In order to solve the overfitting problem in FSL, \citet{vinyals2016matching} proposed an effective episodic meta-learning strategy that learns a generic classifier from diverse few-shot classification tasks and then employs the classifier to a new task. The purpose of episodic training is to mimic the real testing environment where tasks contain insufficient support sets and unlabeled query sets. The consistency between training and testing environments alleviates the shift gap and boosts the generalization. Specifically, using the episodic strategy, the whole process of meta-learning can be divided into three parts: \textbf{meta-training} with training tasks
$\{\mathcal{T}_{\mathrm{train}}^k\}_{k=1}^{N_{\text{train}}}\!=\!\{\mathcal{S}^k,\mathcal{Q}^k\}_{k=1}^{N_{\text{train}}}$, \textbf{meta-validation} with validation tasks 
$\{\mathcal{T}_{\mathrm{val}}^k\}_{k=1}^{N_{\text{val}}}\!=\!\{\mathcal{S}^k,\mathcal{Q}^k\}_{k=1}^{N_{\text{val}}}$, and \textbf{meta-testing} with testing tasks $\{\mathcal{T}_{\mathrm{test}}^k\}_{k=1}^{N_{\text{test}}}\!=\!\{\mathcal{S}^k,\mathcal{Q}^k\}_{k=1}^{N_{\text{test}}}$. Note that for meta-training and meta-validation tasks, the label for the query set is available to train the model $G$ and to select best hyper-parameters, respectively. In this way, meta-learning algorithms are capable of adapting to new tasks effectively even with a shortage of training data for each new task.


\begin{figure*}[t]
\centering
\includegraphics[width=1.0\textwidth]{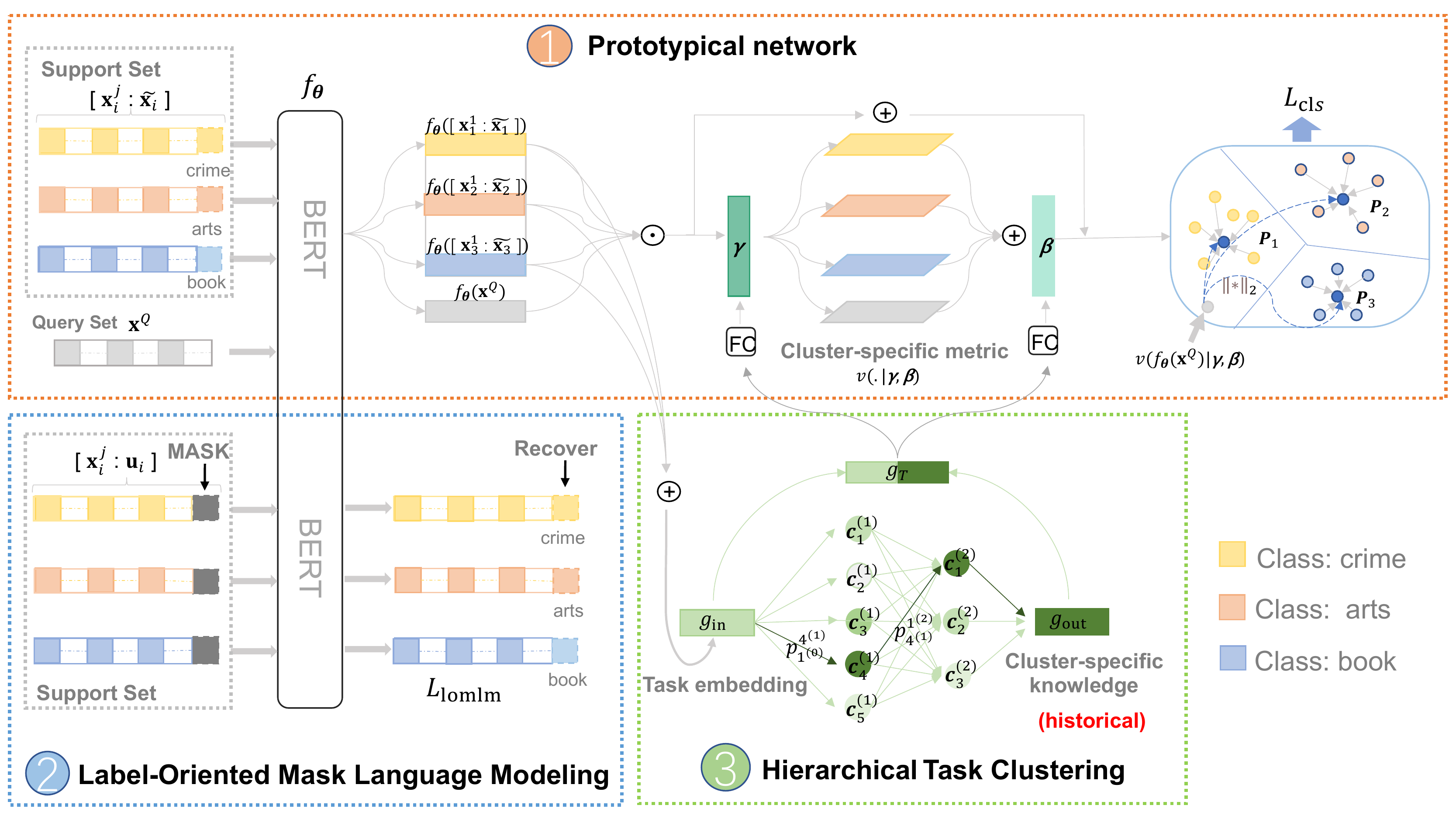}
\vspace{-3mm}
\caption{The SS-HTC framework with a 3-way 1-shot classification task (i.e., crime, arts, book).}
\vspace{-5mm}
\label{fig:framework}
\end{figure*}

\subsection{Problem Formulation}
\noindent \textbf{Multi-source FSTC} \indent
Prior works assume that all the tasks are sampled from a single dataset $D$, making the tasks lying in the same distribution. In this way, we usually split $D$ into three parts: $D_{\textrm{train}}$, $D_{\textrm{dev}}$, and $D_{\textrm{test}}$ in terms of class splits. Each part has a specific label space and disjoints with other parts. For each training episode, we first sample a label set $C$ with $N$ classes from $D_{\textrm{train}}$, and then use $C$ to sample a task $\mathcal{T}_{\mathrm{train}}^{k}$ containing the support set $\mathcal{S}$ and the query set $\mathcal{Q}$. Finally, we feed $\mathcal{S}$ and $\mathcal{Q}$ to the model and minimize the loss. 

This assumption restricts the task diversity and degrades the model's out-of-distribution generalization. To resolve it, we assume that the tasks can be sampled from $M$ diverse datasets $\{D^{1},\ldots,D^{M}\}$ with possibly different distributions. For each dataset $D^{m}$, we use the same strategy to split $D^{m}$ into the training, validation, and testing parts. And we sample 
the meta-training, meta-validation, and meta-testing tasks based on the corresponding parts of all the datasets. That is, we sample $\{\mathcal{T}_{\textrm{train}}\}$ from $\small{\{D^{1}_{\textrm{train}}\!\cup\! D^{2}_{\textrm{train}}\!\cup\!...D^{M}_{\textrm{train}}\}}$, while each task is sampled to consist of only classes from a single dataset.

\section{Method}
SS-HTC aims to handle the task heterogeneity by automatically organizing tasks into a hierarchical task structure that explicitly tailors the transferable knowledge to different task clusters.The overall framework of SS-HTC is illustrated in Figure~\ref{fig:framework}.

SS-HTC mainly consists of three components:
\begin{itemize}
    \item \textbf{Prototypical network (ProtoNet)}~\cite{snell2017prototypical} is an advanced metric-based model, which learns to predict by comparing the distance between the labeled support and unlabeled query sets. We choose it as the building block (\textbf{Base Model}) since it is computationally efficient and simple. More importantly, our framework is general and can be easily compatible with any other metric-based models, e.g., Matching Network~\cite{vinyals2016matching} and Signature~\cite{bao2020fewshot}.
    \item  \textbf{Label-Oriented Mask Language Modeling (LOMLM)} is a self-supervised learning objective to automatically learn the task embedding of each few-shot task $\mathcal{T}$ by considering informative label text semantics. LOMLM encourages the model to generate discriminative task embeddings, which are the prerequisite to identify underlying task relationships for knowledge organization and reuse.
    \item  \textbf{Hierarchical Task Clustering (HTC)} can automatically  group task knowledge into a hierarchical clustering tree, by softly assigning highly correlated tasks into the same cluster, while keeping irrelevant tasks apart. When a new task arrives, it can leverage the historical knowledge within the clusters it belongs to customize a cluster-specific metric for the prototypical network.
\end{itemize} 

\subsection{Prototypical Network}
The prototypical network~\cite{snell2017prototypical} is a simple yet effective metric-based method that learns to predict the label of a query sample $\mathbf{x}^{\mathcal{Q}}$ by comparing its distance with each class prototype vector. Specifically, given a $N$-way $K$-shot task $\mathcal{T}$ defined in Section~\ref{sec:fsl_definition}, we use a prototype vector $\mathbf{p}_{i}$ as the representative feature of each class $y_{i}$, where ${\mathbf{p}_{i}}$ is the average of all the embedded support samples $\{{  \mathbf{x}  }_{i}^{j}\}_{j=1}^{K}$ that belong to class $y_{i}$, i.e., $\small{\mathbf{p}_{i}\!=\!\frac { 1 }{ K } \sum_{j=1}^{K}f_{\bm \theta}({  \mathbf{x}  }_{i}^{j})}$, where $f_{\bm \theta}({  \mathbf{x}  })$ denotes the embedding of a sample. 
Here, we use the pre-trained language model BERT~\cite{devlin2019bert} as the powerful encoder $f_{\bm \theta}$. Then the probability distribution over the $N$ classes for the query sample $\mathbf{x}^{\mathcal{Q}}$ can be calculated via a softmax function over distances between all the prototype vectors and the embedding for $\mathbf{x}^{\mathcal{Q}}$ as
\begin{equation}  \small
\hat{ y }^{ \mathcal{Q} }\!=\!\frac { {\exp}(- d(f_{\bm \theta}({  \mathbf{x}  }^{\mathcal{Q}}), \mathbf{p}_{i})) }{ \sum _{ i'=1 }^{ N }{ {\exp}(- d(f_{\bm \theta}({  \mathbf{x}  }^{\mathcal{Q}})  , \mathbf{p}_{i'}))  }  },
\label{eq:proto}
\end{equation}
where $d(\cdot,\cdot)$ denotes the Euclidean distance. The training objective is to minimize the $N$-way cross-entropy loss $\ell$ for all the query samples in the query set $\mathcal{Q}$ for each meta-training task as
\begin{equation*}  \small
\mathcal{L}_{\mathrm{cls}}\!=\!\sum_{\mathcal{Q}}\ell(y^{ \mathcal{Q} }, \hat{ y }^{ \mathcal{Q} }).
\vspace{-3mm}
\end{equation*}
However, the prototypical network relies on a globally shared metric $(d, f_{\bm \theta})$, which may lack the ability to handle heterogeneous tasks lying in different distributions. Thus, the proposed SS-HTC method uses it as the base model and aims to improve it with the cluster-specific metric to tackle the task heterogeneity problem. 

\subsection{Label-Oriented Mask Language Modeling}
Learning a superior task embedding is the prerequisite to capture underlying correlations between tasks. On the one hand, prior works rely on learning intermediate hidden representations and then aggregate hidden representations of all the training samples as the task embedding~\cite{zamir2018taskonomy}. This may be infeasible in the few-shot regime since the labeled training data $\mathcal{S}$ (i.e., $\small{N\times K}$ examples) of each few-shot task $\mathcal{T}$ are insufficient. On the other hand, existing methods for FSTC~\cite{gao2019hybrid,geng2019induction,sun2019hierarchical,bao2020fewshot} only treat each task as a simple $N$-way classification by mapping informative class label names of each task into indices $\{\small{0,1,...,N\!-\!1}\}$. As such, the model can only focus on discriminating among classes instead of realizing what categories to be classified. Thus, each task is actually underrepresented due to ignoring label semantics.

Inspired by those, we propose a {Label-Oriented Mask Language Modeling (LOMLM)} that exploits underused label semantics to enhance the task representation learning. The LOMLM uses the same denoising auto-encoding~\cite{devlin2019bert} from BERT as the self-supervised learning objective. Specifically, we augment each support text sample ${\mathbf{x} }_{ i }^{j }$ with the label name tokens $\widetilde{\mathbf{x}}_{i}$ of its corresponding class $y_{i}$ (e.g., $\widetilde{\mathbf{x}}_{i}$ is ``{\it musical instruments}'' for the class $y_{i}\!=\!0$). We denote the augmented support sample as $[{\mathbf{x} }_{ i }^{j }; \widetilde{\mathbf{x}}_{i}]$. Then we mask each token in the label name with a special symbol $\mathrm{[MASK]}$, and use the remaining tokens to recover them.\footnote{We explain no information leakage in Appendix~\ref{appendix:leakage}.} 

Let the masked tokens $\{\mathrm{[MASK]}_{t}\}_{t=1}^{T=|\widetilde{\mathbf{x}}_{i}|}$ be ${\mathbf{u}}_{i}$. Then, the training objective of LOMLM is to reconstruct $[{\mathbf{x} }_{ i }^{j }; \widetilde{\mathbf{x}}_{i}]$ from $[{\mathbf{x} }_{ i }^{j }; {\mathbf{u}}_{i}]$ over all support samples by minimizing $\mathcal{L}_{\mathrm{lomlm}}$, which is formulated as
\vspace{-3mm}
\begin{equation*} \small
\mathcal{L}_{\mathrm{lomlm}}\!=\!-\sum_{i=1}^{N}\sum_{j=1}^K\log P([{\mathbf{x} }_{ i }^{j }; \widetilde{\mathbf{x}}_{i}]  |  [{\mathbf{x} }_{ i }^{j }; {\mathbf{u}}_{i}]).
\vspace{-2mm}
\end{equation*}

Under the LOMLM supervision, we simply use an average pooling to aggregate the embeddings of all augmented support samples $[{\mathbf{x} }_{ i }^{j }; \widetilde{\mathbf{x}}_{i}]$ as the representation of the task $\mathcal{T}$ by
\begin{equation}
\label{eq:task_emb}
\mathbf{g}_{\mathrm{in}}=\mathrm{Pool}_{i,j=1}^{N,K}(f_{\bm \theta}([{\mathbf{x} }_{ i }^{j }; \widetilde{\mathbf{x}}_{i}])).
\end{equation}

\subsection{Hierarchical Task Clustering}

To cluster tasks into different groups, where the knowledge from similar historical tasks can be accumulated together and transferred to newly related tasks, we propose a hierarchical task clustering (HTC) to dynamically locate which cluster the task belongs to. The hierarchical structure adopts the top-down hierarchy design to imitate the product taxonomy from coarse to fine granularity (e.g., the ``{\it electronics}'' category has more specific sub-categories such as ``{\it laptop}'', ``{\it phone}'', and ``{\it TV}''). Given complex dependencies among tasks, hierarchical levels of task clusters are more sufficient to capture real-world task relations than the flat clustering ~\cite{kim2010tree}. This allows the task organization and reuse in a coarse-to-fine manner, which can better disentangle inherent task relations such that transferable knowledge among tasks can be maximally leveraged.

In the hierarchical cluster tree, each task $\mathcal{T}$ is soft-assigned into the clusters in each level to encourage less information loss compared with the hard assignment and allow SS-HTC to be trained in an end-to-end manner. Specifically, the assignment score for the next level is a function of the task embedding at the current level. For example, we assign the task embedding $\mathbf{g}^{(l)}_{o}$ in the $o$-th cluster of the $(l)$-th level to the $o'$-th cluster of the $(l\!+\!1)$-th level with the probability $p_{o^{(l)}}^{o'^{(l+1)}}$, which is computed by applying the softmax function over Euclidean distances between $\mathbf{g}^{(l)}_{o}$ and all the $(l\!+\!1)$-th level cluster centers $\{\mathbf{c}^{(l+1)}_{o'}\}_{o'=1}^{O^{(l+1)}}$
\vspace{-3mm}
\begin{equation*}
\small{
p_{o^{(l)}}^{o'^{(l+1)}}\!=\!\frac{\exp(-||\mathbf{g}^{(l)}_{o}-\mathbf{c}_{o'}^{(l+1)}||_{2}^{2}/2{ \sigma}^{2})}{\sum_{o'=1}^{O^{(l+1)}}\exp(-||\mathbf{g}^{(l)}_{o}-\mathbf{c}_{o'}^{(l+1)}||_{2}^{2}/2{ \sigma}^{2})} 
},
\vspace{-2mm}
\end{equation*}
where ${ \sigma}^{2}$ is a scaling factor to control the distance between tasks and clusters and $O^{(l+1)}$ denotes the number of clusters in the $(l\!+\!1)$-th level. Then, the task embedding $\mathbf{g}^{(l+1)}_{o'}$ of the $o'$-th cluster in the $(l\!+\!1)$-th level can be calculated by the weighted sum of all the task embeddings in the previous $l$-th level as
\vspace{-3mm}
\begin{equation*}
\small{
\mathbf{g}^{(l+1)}_{o'}={\sum_{o=1}^{O^{(l)}}p_{o^{(l)}}^{o'^{(l+1)}}\tanh(\textbf{W}_{o'}^{(l+1)}\mathbf{g}_{o}^{(l)}+\textbf{b}_{o'}^{(l+1)})} ,
}
\vspace{-2mm}
\end{equation*}
where $\small{\textbf{W}_{o'}^{(l+1)}}$ and $\textbf{b}_{o'}^{(l+1)}$ are learnable parameters.
The full pipeline of HTC starts from $l\!=\!0$ and $O^{(l)}$ = 1, where the initialization for $\mathbf{g}^{(0)}_{1}$ is the input task embedding $\mathbf{g}_{\mathrm{in}}$ defined in Eq. (\ref{eq:task_emb}), and ends at $O^{(L)}\!=\!1$. The output embedding $\mathbf{g}_{\mathrm{out}}\!=\!\mathbf{g}_{1}^{(L)}$ from the tree encrypts the cluster-specific historical knowledge that can be transferred to the input task. {\it Note that we provide more details for the working mechanism of HTC in our Appendix~\ref{appendix:htc}}.


\noindent\textbf{Cluster-specific feature transformation} \indent After obtaining the cluster-specific knowledge $\mathbf{g}_{\mathrm{out}}$ from the tree that is highly correlative and transferable to the task, we concatenate the input and output task embeddings for the tree as the final task embedding, i.e., $\mathbf{g}_{\mathcal{T}}=\mathbf{g}_{\mathrm{in}}\oplus\mathbf{g}_{\mathrm{out}}$. The task embedding $\mathbf{g}_{\mathcal{T}}$ is used to learn the cluster-specific feature transformation $v(\cdot|{\bm \gamma}, {\bm \beta})$ for the augmented support samples and query samples, which consists of two factors ${\bm \gamma}$ and ${\bm \beta}$ both derived from $\mathbf{g}_{\mathcal{T}}$ as 
\begin{equation*}\small
\begin{split}
{\bm \gamma} &=\rho(\mathbf{ W }_{ \gamma  }\mathbf{g}_{\mathcal{T}}+\mathbf{ b }_{ \gamma  }), \\
{\bm \beta} &=\rho(\mathbf{ W }_{ \bm \beta  }\mathbf{g}_{\mathcal{T}}+\mathbf{ b }_{ \beta}),
\end{split}
\end{equation*}
where $\rho$ denote the ReLU function. ${\bm \gamma}$ and ${\bm \beta}$ are learnable scaling and shift parameters of the feature-wise transformation, which can dynamically adjust feature representations to be more discriminative based on the cluster-specific task embeddings such that it can well adapt to diverse task distributions. Recall that $f_{\bm \theta}(\mathbf{x})$ is the BERT representation of a sample $\mathbf{x}$, where $\mathbf{x}$ can be an augmented support sample $[{\mathbf{x} }_{ i }^{j }; \widetilde{\mathbf{x}}_{i}]$ or a query sample $\mathbf{x}^{\mathcal{Q}}$. For simplicity, let $\mathbf{h}=f_{\bm \theta}(\mathbf{x})$, then the two factors will make a residual affine transformation $v(\cdot|{\bm \gamma}, {\bm \beta})$ on $\mathbf{h}$ as
\begin{equation*}\small
v(\mathbf{h}|\bm \gamma, \bm \beta)=\rho((\mathbf{1}+{\bm \gamma} ) \odot \mathbf{h} + {\bm \beta} ) + \mathbf{h},
\end{equation*}
where $\odot$ is the element-wise multiplication. With the aid of the proposed SS-HTC, we will use the cluster-specific transformed embeddings $v(f_{\bm \theta}(\mathbf{x})|\bm \gamma, \bm \beta)$ instead of the BERT feature embedding $f_{\bm \theta}(\mathbf{x})$ for the inference of the prototypical network as defined in Eq. (\ref{eq:proto}).


\subsection{Joint Training}
We combine each component loss into an overall object function as
\vspace{-2mm}
\begin{equation*}\small
\mathcal{L}=\mathcal{L}_{\mathrm{cls}}+\lambda\mathcal{L}_{\mathrm{lomlm}},
\vspace{-3mm}
\end{equation*}
where $\lambda$ is a hyper-parameter to balance the classification loss and the LOMLM loss. The goal of joint learning is to learn superior task embeddings to guide the cluster-specific discriminative learning for the ultimate few-shot text classification.

\section{Experiments}
\subsection{Setup}
\noindent \textbf{Datasets} \indent
We evaluate SS-HTC on five FSTC benchmark datasets: {Amazon Product Review}~\cite{he2016ups}, {20 Newsgroups}~\cite{lang1995newsweeder}, {HuffPost}~\cite{misra2018news}, {Reuters}~\cite{lewis1997reuters}, and {RCV1}~\cite{lewis2004rcv1}. Following~\cite{bao2020fewshot}, we use the same class splits to divide each dataset into {meta-training}, {meta-validation} and {meta-testing} parts, from which $N$-way $K$-shot tasks are randomly sampled. 

\noindent \textbf{Setting} \indent
In experiments, tasks can be sampled from multiple diverse datasets. Thus, all models are trained and evaluated on the combination of the five aforementioned benchmark datasets instead of each dataset separately. Following prior works on FSTC, the classification accuracy is used as the evaluation metric as each task is under the few-shot learning setting and has no data imbalance issue. Moreover, we use the average accuracy on randomly sampled $1,000$ meta-testing tasks for each dataset as the final results to avoid the problem of randomness. All the experiments repeat 3 times and average results over 3 runs are reported.

\begin{table*}[t]
\centering
\resizebox{2.0\columnwidth}{!}{
\begin{tabular}{ccc|cc|cc|cc|cc|cc}
\Xhline{2\arrayrulewidth}
\multirow{2}{*}{\textbf{Model}} &\multicolumn{2}{c}{20 News} &\multicolumn{2}{c}{Amazon} &\multicolumn{2}{c}{Huffpost} &\multicolumn{2}{c}{Reuters}   
&\multicolumn{2}{c}{RCV1} &\multicolumn{2}{c}{\textbf{Avg}}\\  \cline{2-13}
 & 1 shot & 5 shot  & 1 shot & 5 shot & 1 shot & 5 shot  & 1 shot & 5 shot & 1 shot & 5 shot & 1 shot & 5 shot \\ 
\hline  
& \multicolumn{11}{c}{{\textbf{Supervised learning}}} \\ 
BERT (FT) & 28.30 & 34.01 & 34.35 & 43.93 & 23.50 & 28.11 & 37.01 & 51.35 & 30.42 & 39.88 & 30.72 & 39.46\\
\hline
& \multicolumn{11}{c}{{\textbf{ Gradient-based meta learning}}} \\ 
Reptile & 33.78 & 40.23 & 39.86 & 55.01 & 25.69 & 36.20 & 47.55 & 64.56 & 	40.02 & 56.33 & 37.38 & 50.47 \\
PMAML & 34.46 & 40.25 & 38.41 & 53.69 & 26.18 & 35.44 & 46.57 & 65.13 & 38.99 & 57.70 & 36.92 & 50.44 \\ 
\hline
& \multicolumn{11}{c}{{\textbf{ Metric-based meta learning}}} \\ 
MatchNet & 38.03 & 40.42 & 39.26 & 34.02 & 33.16 & 58.02 & 54.27 & 38.16 & 40.61 & 42.44 & 41.07 & 42.61\\
InductionNet & 41.35 & 43.52 & 46.36 & 43.18 & 38.09 & 42.32 & 69.12 & 66.92 & 46.04 & 50.84 & 48.19 & 49.36\\
ProtoNet & 49.30 & 65.51 & 68.91 & 84.79 & 46.54 & 65.55 & 73.46 & 87.45 & 49.32 & 69.70 & 57.51 & 74.60\\
HybridAPN & 48.56 & 57.80 & 68.92 & 77.25 & 44.39 & 53.19 & 80.21 & 88.42 & 55.95 & 66.52 & 59.61 & 68.64 \\
HierAPN & 52.88 & 57.66 & 66.35 & 75.64 & 41.87 & 51.62 & 80.28 & 92.58 & 54.83 & 61.54 & 59.24 & 67.81\\
Signature & 52.48 &	66.20 & 66.64 & 84.44 & 45.32 & 63.20 & 83.52 &	93.20 & 	53.58 & 69.20 &	60.31 & 75.25\\
DEM & 49.88 & 57.93 & 54.96 & 73.56 & 52.34 & 69.66 & 87.27 &	95.21 & 58.18 & 76.07 &	60.53 & 74.49\\
\hline
\textbf{SS-HTC} & \textbf{58.77}$^{\dag}$ & \textbf{69.24}$^{\dag}$  & \textbf{75.92}$^{\dag}$ & \textbf{86.84}$^{\dag}$ & \textbf{63.72}$^{\dag}$ & \textbf{71.88}$^{\dag}$ & \textbf{89.36}$^{\dag}$ & \textbf{95.98}$^{\dag}$ & \textbf{63.91}$^{\dag}$  & \textbf{78.24}$^{\dag}$& \textbf{70.34}$^{\dag}$ & \textbf{80.44}$^{\dag}$ \\
$\Delta$ & (+6.29) & (+3.04) & (+8.65) & (+2.40) & (+11.38) & (+2.22) & (+2.09) & (+0.77) & (+5.73) & (+2.17) & (+9.81) & (+5.19) \\
\Xhline{1.5\arrayrulewidth}
\end{tabular}
}
\caption{Main results: 5-way $K$-shot evaluation. $\Delta$ refers to the improvements over the best baseline. $^{\dag}$ means the statistically significant improvement with paired sample $t$-test with $p$-value $< 0.01$.}
\label{table:k_shot}
\vspace{-0.25in}
\end{table*}

\noindent \textbf{Baselines} \indent
\noindent  For a fair comparison, we use the BERT as the base encoder for all baselines. 

\noindent $\bullet$ \textbf{Supervised learning.} \noindent \textbf{BERT (FT)}~\cite{chen2019closerfewshot} trains a BERT with a generic $N$-way classifier on all meta-training tasks and finetunes it on the support set and evaluate it on the query set of each meta-testing task independently.

\noindent $\bullet$ \textbf{Gradient-based meta-learning} methods aim to learn a well-generalized model initialization that can be adapted to new tasks within a few optimization steps. \noindent (\romannumeral1) \textbf{Reptile}~\cite{nichol2018first} is a fast first-order gradient approximation of MAML which could be hardly optimized based on BERT~\cite{finn2017model}. (\romannumeral2) \textbf{PMAML}~\cite{2019Improvingpmaml} employs the masked language model pretraining before using the first-order MAML.

\noindent $\bullet$ \textbf{Metric-based meta-learning} methods are to learn an invariant metric space where classes can be differentiated between each other. (\romannumeral1) \textbf{MatchNet}~\cite{vinyals2016matching} uses an attention-based scheme where the cosine distance is used as the metric. (\romannumeral2) \textbf{ProtoNet}~\cite{snell2017prototypical} learns a metric space by minimizing the Euclidean distance between class prototype and query samples. (\romannumeral3) \textbf{InductionNet}~\cite{geng2019induction} encapsulates different classes by a dynamic routing induction method. (\romannumeral4) \textbf{HybridAPN}~\cite{gao2019hybrid} is a hybrid attention prototypical network that exploits a hybrid attention mechanism. (\romannumeral5) \textbf{HierAPN}~\cite{sun2019hierarchical} is a hierarchical attention prototypical network that designs a hierarchical attention mechanism. (\romannumeral6) \textbf{Signature}~\cite{bao2020fewshot} utilizes the distributional statistics to implement the attention transfer between tasks. (\romannumeral7) \textbf{DEM}~\cite{DEMfewshot} introduces a difference extractor to derive distinctive label representations with multi-task learning based on ProtoNet.

\subsection{Implementation details}
\noindent \textbf{Environment} \indent Our proposed SS-HTC model and baseline methods are implemented in TensorFlow 2.4.0 with CUDA 10.1, using Python 3.7.0 from Anaconda 4.9.2. All the models are trained/tested on a single TESLA V100-PCIE 32GB GPU with Linux system.

\noindent \textbf{Encoder} \indent  We use the BERT-base model: {\it bert-base-uncased}~\cite{wolf2019huggingface}
model as the encoder, which has 12 layers, 768-dimensional hidden representations, 12 heads, and 110M parameters in total. We use the pooled representation (i.e., averaged token embeddings) as the sentence embedding since we have found that [CLS] embedding performs very poorly, even worse than CNN encoders under the few-shot setting. The BERT is jointly optimized with other parameters during the training stage. 
 
\noindent \textbf{Initialization \& Training} \indent  For all the experiments, SS-HTC is optimized by the Adam algorithm~\citep{kingma2014adam} for training. The maximal sentence length is 450. The weight matrices are initialized with a uniform distribution $\small{U( -0.01,0.01 )}$. Gradients with the $l2$ norm larger than 40 are normalized to be 40. To alleviate overfitting, we perform early stopping on the meta-validation tasks. 

\noindent \textbf{Hyperparameter} \indent  The hyper-parameters are manually tuned on the average accuracy of the 10\% randomly held-out meta-training sets. The initial learning rate is $10^{-5}$, which is tuned amongst \{$10^{-6}$, $5\times10^{-6}$, 1$\times10^{-5}$, $5\times10^{-5}$\}. The weight ${\lambda}$ for $\mathcal{L}_{\mathrm{lomlm}}$ is 0.1, which is tuned amongst \{0.01, 0.03, 0.1, 0.3\}. The scaling factor $\sigma^2$ is 2.0. Due to the limited GPU memory, we only feed one task to SS-HTC for each step.

\subsection{Main Results}
\textbf{$K$-Shot Evaluation}. We present in Table~\ref{table:k_shot} experimental results in terms of different shots under the setting of five ways/classes. Based on the results, we can observe:
\textbf{SS-HTC}: SS-HTC significantly and consistently outperforms all the baseline methods on five datasets by a large margin (i.e., {1-shot:+9.81\%}, {5-shot:+5.19\%} average accuracy) over the best baselines (i.e., Signature and DEM).

\noindent $\bullet$ \textbf{Supervised method}: Even with powerful pre-trained language models (PLMs) like BERT, the supervised method BERT (FT) still performs very poorly in the few-shot regime. This circumstance has also been shown in prior studies~\cite{yogatama2019learning}, which shows that PLMs highly rely on sufficient fine-tuning data for downstream tasks.

\noindent $\bullet$ \textbf{Gradient-based methods}: As gradient-based methods, Reptile and PMAML show inferior performance to metric-based baselines in FSTC. This phenomenon has also been observed in recent FSTC works~\cite{gao2019hybrid,bao2020fewshot}. The gradient-based methods mainly focus on low-noise vision tasks, which makes them hard to directly deal with diverse and noisy text data in FSTC tasks, especially for our setting that tasks are coming from multiple resources with large diversity.

\noindent $\bullet$ \textbf{Metric-based methods}: (\romannumeral1) Compared with gradient-based methods,  most metric-based baselines can generally obtain better results for FSTC. (\romannumeral2) Recent proposed text-specific metric-based methods like HybridAPN and HierAPN have better performance than their base model - ProtoNet when tasks are all sampled from a single dataset~\cite{gao2019hybrid,sun2019hierarchical}. However, when tasks are heterogeneous from diverse datasets in our setting, they do not outperform ProtoNet. This indicates that their sophisticated metric designs may not be able to handle the task heterogeneity due to the global knowledge-sharing strategies used. (\romannumeral3) SS-HTC can outperform those metric-based baselines. This is because that SS-HTC can customize the transferable knowledge to be cluster-specific and preserve knowledge generalization among highly related tasks by taking advantage of the dynamic task clustering.

\begin{table*}[t]
\begin{tabular}{cc}
\includegraphics[width=3.in]{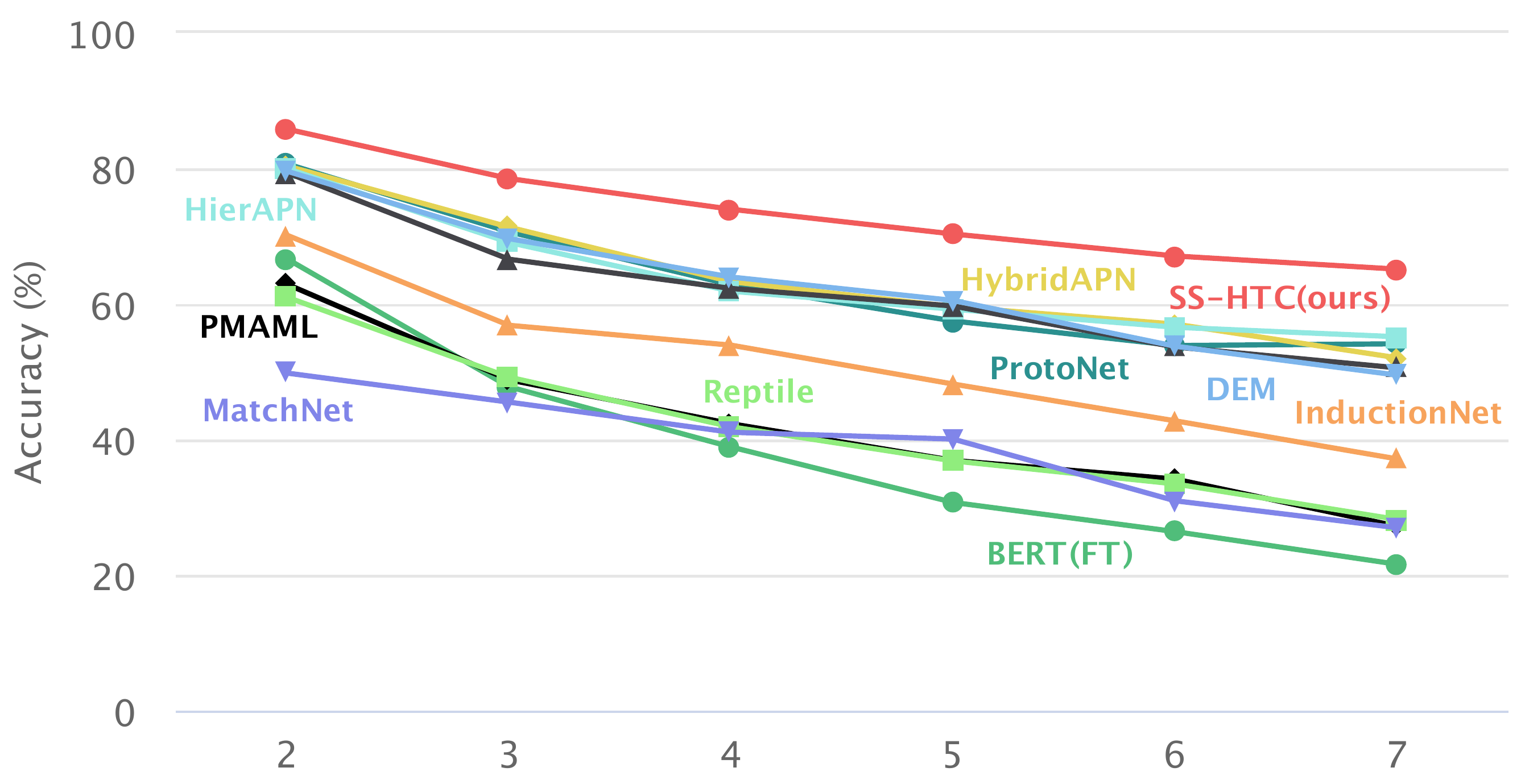} &\includegraphics[width=3.in]{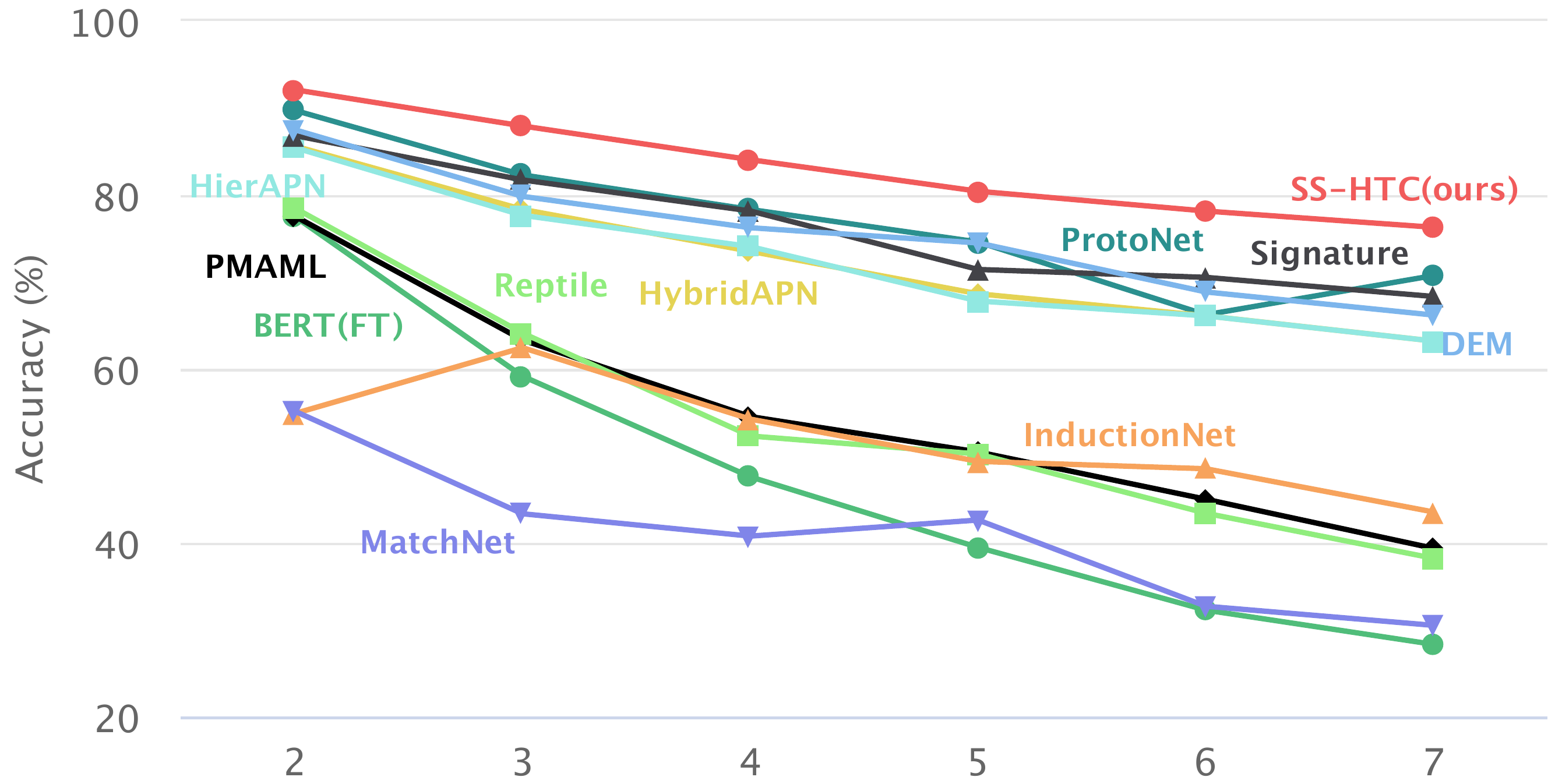} \\
\multicolumn{2}{c}{\includegraphics[width=6.0in]{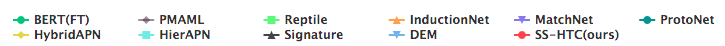}} \\
 (a) \textbf{$\mathbf{N}$-way 1-shot} & (b) \textbf{$\mathbf{N}$-way 5-shot} \\
 \end{tabular}
\vspace{-2mm}
\captionof{figure}{Main results: Average performance for $N$-way evaluation with the fixed numbers of shots.}
\vspace{-2mm}
\label{fig:N_way}
\end{table*}

\begin{table}[t]
\centering
\resizebox{0.9\columnwidth}{!}{
\begin{tabular}{lcc|cc}
\Xhline{1.5\arrayrulewidth}
{Model}  & 1-shot & 5-shot &{{Avg}}  & Gain  \\ \hline \hline  
{ProtoNet} &  57.51 & 74.60 & 66.06  & -\\
{ProtoNet+HTC} & 60.98 & 76.04 & 68.51 & + 2.45\\  \hline
\multicolumn{2}{l}{ProtoNet+LOMLM+HTC} \\ 
- a.k.a SS-HTC  & \textbf{70.34} & \textbf{80.44} & \textbf{75.39} & +6.88 \\ 
\Xhline{1.5\arrayrulewidth}
\end{tabular}}
\caption{Average results for 5-way classification.}
\label{table:ablation}
\vspace{-6mm}
\end{table}

\noindent \textbf{$N$-way Evaluation}. We present in Figure~\ref{fig:N_way} the results in terms of different ways with a fixed number of shots. We report the average accuracy across all the five datasets with $\small{N\!=\!{2,3,...,7}}$. Generally, as the number of ways increases, the performance degrades as the FSTC tasks become more difficult. We can observe that SS-HTC performs better than other baselines and that the gap among them becomes larger as the number of ways increases. This indicates the proposed SS-HTC method is less sensitive to the difficulty of the FSTC task by leveraging the knowledge from the most similar tasks based on hierarchical task clustering.

\subsection{Ablation Study}
To verify the efficacy of each component, we progressively incorporate the hierarchical task clustering (HTC) and label-oriented masked language modeling (LOMLM) into the base model (i.e., ProtoNet). We present the ablation results in Table~\ref{table:ablation}.

\noindent $\bullet$ \textbf{w/ HTC} v.s. \textbf{w/o HTC}: For ProtoNet+HTC, we use the proposed HTC method to dynamically organize tasks into hierarchical clusters, where the knowledge from similar tasks can be accumulated together. As such, each new incoming task can leverage the transferable knowledge within the cluster it belongs to and customize the cluster-specific metric for few-shot learning. We observe HTC can bring a significant gain (i.e., 2.45\%) over ProtoNet in terms of the average accuracy. This shows the effectiveness of HTC to handle heterogeneous tasks lying in different distributions. 

\noindent $\bullet$ \textbf{w/ LOMLM} v.s. \textbf{w/o LOMLM}: For ProtoNet+HTC, we simply average embeddings of all support samples and their corresponding label texts as the embedding of a task without any supervision. Thus, this task embedding could be underrepresented. By incorporating the LOMLM, the task embedding is enhanced to be more label-aware to discriminate among classes. According to Table~\ref{table:ablation}, we observe that adding LOMLM can achieve an additional 6.88\% gain in terms of the average accuracy, which is a very large improvement over ProtoNet+HTC. This implies that a superior task embedding is critical to better disentangling task relations and customize the cluster-specific metric.

\subsection{The Effect of Tree Structure}
We further study the effect of tree structure to the performance. We vary the tree structure and record the results in Table~\ref{table:cluster}. From the results, we can observe that the proposed hierarchical clustering shows the superiority over the flat task clustering. For the hierarchical clustering, we can see that too few clusters may be insufficient to learn the task clustering characteristic (e.g., the case (2,2,1)). When we increase the number of clusters, SS-HTC can achieve better results (e.g., case (5,3,1)) until reaching a stable status (e.g., case (5,5,1)). This indicates that more clusters introduce more parameters and may result in the overfitting problem.

\begin{table}[htb]
\centering
\resizebox{0.65\columnwidth}{!}{
\begin{tabular}{lcc|cc}
\Xhline{2\arrayrulewidth}
Num. of Clu.  & 1-shot & 5-shot &{Avg}   \\ \hline \hline \multicolumn{4}{c}{{\textbf{Flat clustering}}} \\ \hline
{(5,1)} & 68.18 & 79.12 & 73.65 \\  
{(15,1)} & 68.41 & 79.41 & 73.91 \\  
\multicolumn{4}{c}{{\textbf{Hierarchical clustering}}} \\ \hline
{(2,2,1)} & 68.14 & 79.20 & 73.67 \\ 
{(3,2,1)} & 68.03 & 79.51  & 73.77 \\ 
{(5,3,1)} & \textbf{70.34} & \textbf{80.44} & \textbf{75.39} \\ 
{(5,4,1)} & 70.12 & 80.18 & 75.15 \\  
{(5,5,1)} & 70.30 & 80.38 & 75.34 \\  
\hline
\Xhline{1.5\arrayrulewidth}
\end{tabular}}
\caption{
Comparison among different cluster \#. $(\cdot,\cdot,\cdot)$ denotes the cluster \# from the bottom to the top layer. Average accuracy for 5-way classification is reported.}
\label{table:cluster}
\vspace{-5mm}
\end{table}

\begin{table*}[t]
\resizebox{0.92\linewidth}{!}{
\begin{tabular}{cc}
\includegraphics[width=3.4in]{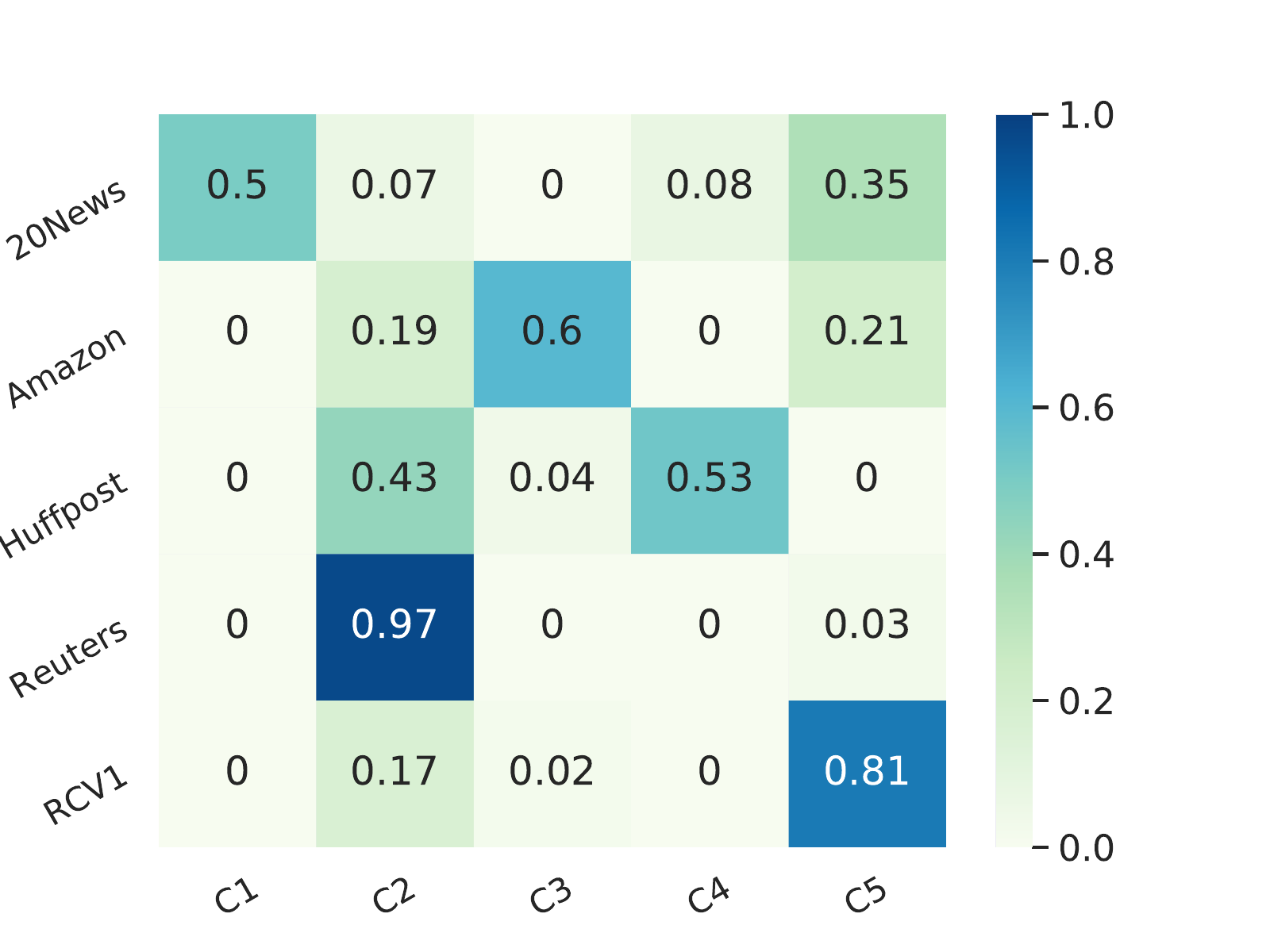} &
\includegraphics[width=4.0in]{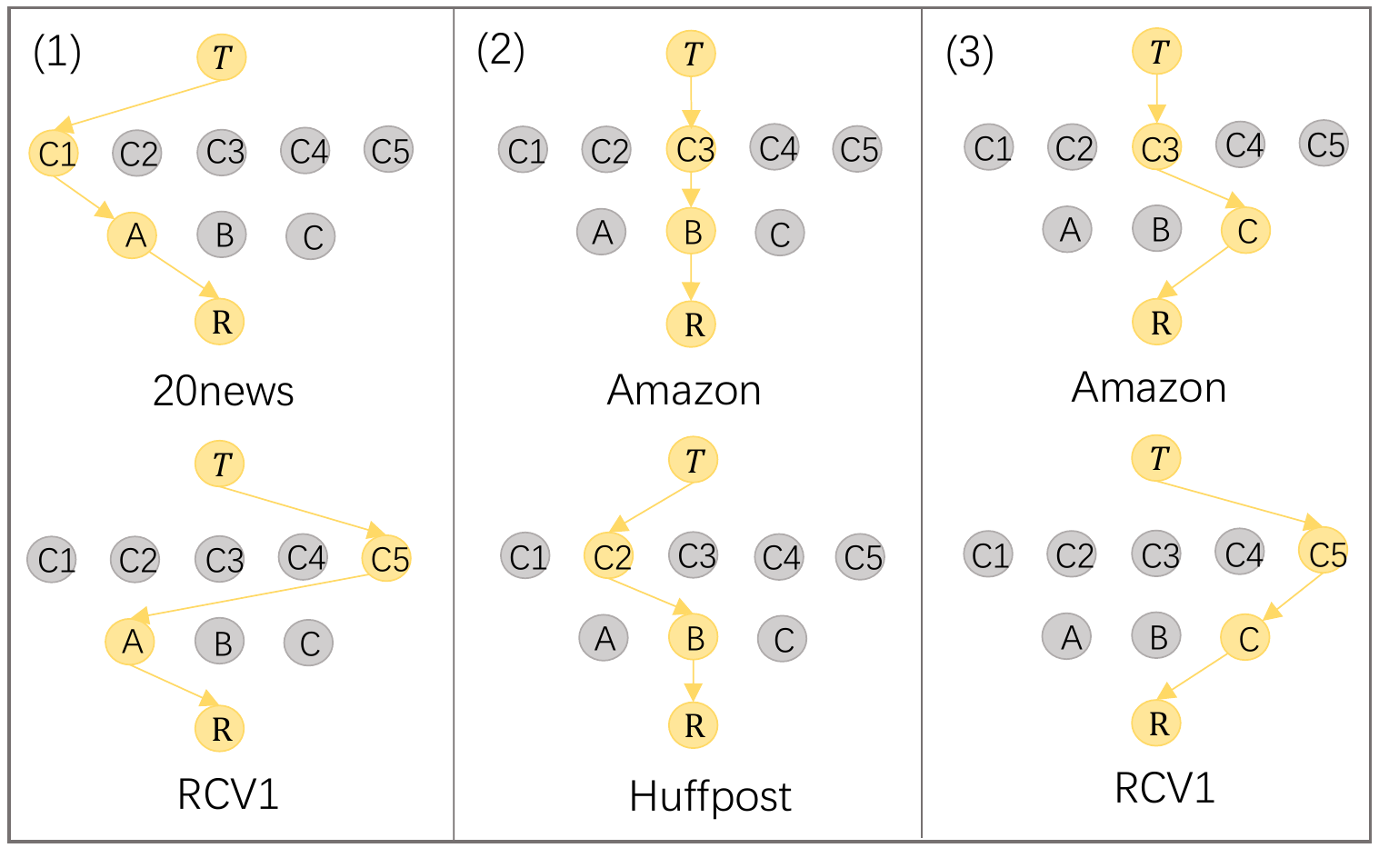} \\
 \end{tabular}}
\vskip -0.1in
\captionof{figure}{Left: visualization of average soft-assignment $p_{o^{(l)}}^{o'^{(l+1)}}$ of 1000 random tasks for each dataset. Right:  hierarchical structure learned from different tasks. The most activated cluster is marked in yellow.}
\vspace{-3mm}
\label{fig:tree_vis}
\end{table*}

\subsection{Visualization of Hierarchical Task Tree}
To demonstrate that the proposed SS-HTC method can automatically disentangle the underlying task relationship, we visualize the SS-HTC with cluster structure $(5,3,1)$ for tasks from each dataset. Specifically, we first select 1,000 5-way 1-shot tasks randomly from each dataset and show their averaged soft-assignments of clusters (C1, C2, C3, C4, C5) in the first layer. As illustrated in the left subfigure in Figure~\ref{fig:tree_vis} where a darker color means a higher probability, we can see that different datasets mainly activate different clusters: Reuters$\rightarrow$C2, Amazon$\rightarrow$C3, RCV1$\rightarrow$C4, and 20News$\rightarrow$C1. Particularly, Huffpost activates both C2 and C4, which indicates that the Huffpost and Reuters datasets may have a large overlap. By checking the classes sets of both datasets, we have found that several classes in the two datasets are highly-related (e.g., Huffpost: ``{\it taste}'' and ``{\it word news}'', Reuters: ``{\it sugar}'' and ``{\it wholesale price Iindex}''). 

Besides, we also explore the activated task clusters (A, B, C) in the second layer which further accumulates the transferable knowledge among tasks from different datasets. We observe tasks from different datasets that have similar classes are highly aggregated into the same cluster. Meanwhile, tasks from the same dataset that contains different classes can activate different clusters. For example, in the \#1 case of  Figure~\ref{fig:tree_vis} Right, a 5-way task from \textbf{20News} with the class set \{``{\it alt atheism}'', ``{\it soc religion christian}'', ``{\it talk politics guns}'', ``{\it talk politics misc}'', ``{\it talk religion misc}''\} and a 5-way task from \textbf{RCV1} with the class set "\{``{\it religion}'', ``{\it equity markets}'', ``{\it domestic politics}'', ``{\it interbank markets}'', ``{\it money markets}''\}" both activate the first cluster A, since the two tasks are all related to religion and politics. Similarly, in the \#2 case, a 5-way task from \textbf{20News} with the class set \{``{\it books}'', ``{\it clothing shoes jewelry}'', ``{\it electronics}'', ``{\it musical instruments}'', ``{\it tools home improvement}''\} and a 5-way task from \textbf{Huffpost} with the class set \{``{\it arts culture}'', ``{\it good news}'', ``{\it environment}'', ``{\it tech}'', ``{\it style}''\} both activate the second cluster B because they are all about culture and life-related stuffs. In the \#3 case, a 5-way task from \textbf{Amazon} with the class set \{``{\it Books}'', ``{\it kindle store}'', ``{\it movies tv}'', ``{\it office products}'', ``{\it tools home improvement}''"\} and a 5-way task from \textbf{RCV1} with the class set \{``{\it economics}'', ``{\it government finance}'', ``{\it management}'', ``{\it performance}'', ``{\it share listings}''\} are both assigned to the third cluster C as they both concern the economy and education. Those qualitative results indicate that the proposed SS-HTC model can capture the latent relations between diverse tasks to improve the model interpretability.


\section{Related Work}
\noindent  \textbf{Meta-Learning} \indent
Inspired by human beings' ability to transfer knowledge from previous experiences~\cite{pan2009survey,li2017end,li2018hierarchical,li2019exploiting,li2019sal}, meta-learning~\cite{vinyals2016matching,finn2017model} has become the mainstream paradigm to resolve few-shot learning problems. Prior studies mainly focus on low-noise vision tasks~\cite{snell2017prototypical,sung2018learning,nichol2018first,oreshkin2018tadam,liu2019tpn}. Recently, those techniques have been initiated to low-resource NLP problems such as few-shot text classification~\cite{yu2018diverse,wu2019learning,geng2019induction,sun2019hierarchical,geng2020dynamic,bao2020fewshot,wang2021gradtask,DEMfewshot}, relation classification~\cite{han2018fewrel,gao2019hybrid,obamuyide2019model,gao2019fewrel}, machine translation~\cite{gu2018meta}, knowledge graph completion~\cite{SS-AGA,wang2022learning}, and natural language understanding~\cite{dou2019investigating,bansal2020learning,bansal2020self,li2021metats} with minimal supervision. Different from them that globally share the prior knowledge across homogeneous tasks within a single source, SS-HTC can embrace the skills learned from multiple heterogeneous sources to improve the out-of-distribution robustness 
More importantly, they neglect underlying task relations in the low-data regime, which is imperative to automatically organize knowledge from heterogeneous tasks.

\noindent  \textbf{Label-aware Modeling} \indent
To alleviate data scarcity, label-aware methods are recently investigated
~\cite{yin2019benchmarking,puri2019zero,meng2020text,halder2020task} to incorporate label semantics into text representation learning. Yin et al.~\cite{yin2019benchmarking} incorporate label texts into text samples and convert the text classification into a text entailment task. Yu et al.~\cite{meng2020text} propose the category vocabulary, which can be good label supplements for our LOMLM to enrich the label semantics in the future work. However, those methods are less investigated in task adaptation for FSTC. More importantly, our ultimate goal aims to leverage label information to enhance few-shot task representations for discovering and disentangling inherent and complicated task correlations. This can facilitate the knowledge organization and handle heterogeneous new tasks as well as improving the model interpretability.

\section{Conclusion}
In this paper, we propose the self-supervised hierarchical task clustering (SS-HTC) method to tackle the task heterogeneity for FSTC by dynamically organizing the tasks into hierarchical clusters and customize the cluster-specific knowledge. Extensive experiments on various FSTC benchmark datasets quantitatively and qualitatively demonstrate the effectiveness of SS-HTC. In the future, the proposed SS-HTC can be potentially 
generalized to the multilingual few-shot setting~\cite{hu2020xtreme}.

\section{Limitations}
Although we introduce a more realistic and practical problem setting for few-shot learning and verify the proposed SS-HTC method on extensive experiments, there are still some future directions that need further investigation and exploration. Firstly, our proposed setting is supposed to generalize to more heterogeneous NLP tasks under the few-shot regime instead of restricting to text classification. Secondly, how to dynamically adapt the task organization structures like humans in terms of input tasks is still underexploited. We view our works as the start point and will further explore those interesting problems in the future work.

\section*{Acknowledgements}

This work is supported by NSFC general grant under grant no. 62076118 and Shenzhen fundamental research program JCYJ20210324105000003.

\bibliography{emnlp2022}
\bibliographystyle{acl_natbib}

\appendix

\section{Clarification}
In this section, we give more clarification regarding the details of problem setting, framework as well as hierarchical task clustering.

The proposed SS-HTC framework exactly comes from the collective power of self-supervised LOMLM and HTC. By this means, SS-HTC balances between globally shared meta-knowledge and cluster-specific meta-knowledge, where the transferable knowledge can be adapted to different clusters of tasks, while it is still shared among highly correlated tasks within the same cluster. 

\subsection{No Information leakage claim}
\label{appendix:leakage}
There is no information leakage for the proposed Label-Oriented Mask Language Modeling (LOMLM) module. We only utilize masked label tokens for support samples (training set) instead of query samples (testing set) in each task. For meta-learning, the final performance evaluation is based on query samples of each meta-testing task that has disjoint classes with all meta-training tasks.

\subsection{Hierarchical Task Clustering}
\label{appendix:htc}
The characteristics of HTC can be summarized as two aspects: (1) the hierarchical task clusters $\{\mathbf{c}^{(l)}_{o}\}_{o=1}^{O^{(l)}}$ in each $(l)$-th level of the tree are learnable and randomly initialized, which are shared by all tasks. We only need to feed each task embedding into the tree, automatically obtain the soft assignment to each cluster, and output the cluster-specific historical knowledge used for the prototypical network. The structure of the hierarchical tree is predefined since we found that jointly learning with additional structures can bring more challenges into the optimization. Despite that, the cluster representations and their connection weights are jointly learned with other parameters in an online manner to model complex task relationships; (2) hierarchical clustering tree is optimized on the task level instead of the class level, which can capture more enriched task-specific information beyond the class itself. We found this information is particularly useful to handle the diversity of few-shot tasks, as our practical setting allows tasks to be sampled from a diverse range of data sources with possibly different data distributions.

\section{Baselines}
We provide the available open source code for the baseline methods we compare with, including: 
\begin{itemize}
    \item \textbf{Supervised learning}. 
    \begin{itemize}
        \item \textbf{BERT (FT)}~\cite{chen2019closerfewshot}\footnote{\url{https://github.com/wyharveychen/CloserLookFewShot}}
    \end{itemize}
    
    \item \textbf{Gradient-based meta-learning} 
    \begin{itemize}
        \item    \textbf{Reptile}~\cite{nichol2018first}\footnote{\url{https://github.com/openai/supervised-reptile}}
        \item \textbf{PMAML}~\cite{2019Improvingpmaml}\footnote{\url{https://github.com/zxlzr/FewShotNLP}}
    \end{itemize}
    \item \textbf{Metric-based meta-learning}
    \begin{itemize}
        \item \textbf{MatchNet}~\cite{vinyals2016matching}\footnote{\url{https://github.com/gitabcworld/MatchingNetworks}}
        \item \textbf{ProtoNet}~\cite{snell2017prototypical}\footnote{\url{https://github.com/jakesnell/prototypical-networks}}
        \item \textbf{InductionNet}~\cite{geng2019induction}\footnote{\url{https://github.com/YujiaBao/Distributional-Signatures}}
        \item \textbf{HybridAPN}~\cite{gao2019hybrid}\footnote{\url{https://github.com/thunlp/HATT-Proto}}
        \item \textbf{Signature}~\cite{bao2020fewshot}\footnote{\url{https://github.com/YujiaBao/Distributional-Signatures}}
        \item \textbf{DEM}~\cite{DEMfewshot}\footnote{\url{https://github.com/21335732529sky/difference_extractor}}
    \end{itemize}
\end{itemize} 
For \textbf{HierAPN}~\cite{sun2019hierarchical}, we reimplement it according to the original paper since the source code is not publicly available.

\end{document}